\documentclass[letterpaper, 10 pt, conference]{ieeeconf}  % Comment this line out if you need a4paper

\IEEEoverridecommandlockouts                              % This command is only needed if 
\usepackage{graphics} % for pdf, bitmapped graphics files
\usepackage{epsfig} % for postscript graphics files
\usepackage{mathptmx} % assumes new font selection scheme installed
\usepackage{times} % assumes new font selection scheme installed
\usepackage{amsmath} % assumes amsmath package installed
\usepackage{amssymb}  % assumes amsmath package installed
\usepackage{subfigure}
\usepackage{color}
\graphicspath{ {figs/} }
\usepackage{cite}

\PassOptionsToPackage{hyphens}{url}\usepackage{hyperref}

\title{\LARGE \bf
A Comparative Analysis of Contact Models in Trajectory Optimization for Manipulation*
}

\author{Aykut \"{O}zg\"{u}n \"{O}nol, Philip Long, and Ta\c{s}k\i n Pad\i r$^{1}$% <-this % stops a space
\thanks{*This research is supported by the Department of Energy under Award Number DE-EM0004482, by the National Aeronautics and Space Administration under Grant No. NNX16AC48A issued through the Science and Technology Mission Directorate and by the National Science Foundation under Award Nos. 1451427, 1544895, 1649729.}
\thanks{$^{1}$RIVeR Lab, Northeastern University, Boston, MA, USA
        {\tt\small \{onol.a, p.long, t.padir\}@northeastern.edu}}%
}

\begin{document}

\maketitle
\thispagestyle{empty}
\pagestyle{empty}

%%%%%%%%%%%%%%%%%%%%%%%%%%%%%%%%%%%%%%%%%%%%%%%%%%%%%%%%%%%%%%%%%%%%%%%%%%%%%%%%
\begin{abstract}

In this paper, we analyze the effects of contact models on contact-implicit trajectory optimization for manipulation. We consider three different approaches: (1) a contact model that is based on complementarity constraints, (2) a smooth contact model, and our proposed method (3) a variable smooth contact model. We compare these models in simulation in terms of physical accuracy, quality of motions, and computation time. In each case, the optimization process is initialized by setting all torque variables to zero, namely, without a meaningful initial guess. For simulations, we consider a pushing task with varying complexity for a 7 degrees-of-freedom robot arm. Our results demonstrate that the optimization based on the proposed variable smooth contact model provides a good trade-off between the physical fidelity and quality of motions at the cost of increased computation time.

\end{abstract}

%%%%%%%%%%%%%%%%%%%%%%%%%%%%%%%%%%%%%%%%%%%%%%%%%%%%%%%%%%%%%%%%%%%%%%%%%%%%%%%%
\section{INTRODUCTION}
Discovering contact-rich motions for manipulation and locomotion tasks without specifying a contact schedule is a captivating idea. Thus, contact-implicit trajectory optimization attracted many researchers from fields such as robotics, computer graphics, and bio-mechanics. The main idea of this approach is to define an optimization problem with costs and constraints describing a task of interest so that a contact schedule and corresponding forces are found as a result of trajectory optimization.

In this method, selection of the contact model is crucial since it provides optimization with a way to infer contact dynamics. Complementarity constraints that describe rigid-body collisions have been extensively used for this purpose. A time-stepping scheme that uses complementarity constraints to model rigid-body dynamics with inelastic collisions and Coulomb friction was first proposed by Stewart \& Trinkle \cite{stewart1996implicit}. Anitescu \& Potra \cite{anitescu1997formulating} modified this method to handle any configuration and any number of contacts. Based on this time-stepping scheme, Posa et al. \cite{posa2014direct} proposed a direct method for contact-implicit trajectory optimization to prevent the ``combinatorial explosion'' of hybrid models. Indeed, the idea of transcribing a non-smooth trajectory optimization problem with impacts and discontinuities into a nonlinear optimization problem with complementarity constraints was first introduced by Yunt \& Glocker \cite{yunt2005trajectory}. However, in \cite{yunt2005trajectory}, the optimization of control inputs is decoupled from the optimization of states and contact forces; whereas, in \cite{posa2014direct}, they are all optimized simultaneously. Nonetheless, in both methods, contact dynamics is modeled as a complementarity problem. \cite{mastalli2016hierarchical} also solves a complementarity problem to find a contact schedule.

On the other hand, smooth contact models facilitate convergence of gradient-based solvers. For this reason, \cite{todorov2012synthesis,todorov2015ensemble,buchli16_relax,manchester2017variational} use smoother fragments of complementarity constraints; whereas \cite{marcucci2017two,buchli17_discovery,buchli17_mpc} directly define contact forces as smooth functions of distance. In \cite{todorov2012discovery} and \cite{todorov2012manipulation}, contacts are modeled within the cost function by using auxiliary variables. In addition, there are other contact modeling approaches that fall outside of this classification such as the ones in \cite{jain2009optimization} and \cite{zimmermann2015hierarchical}.

Although contact-implicit trajectory optimization is, by definition, independent of task and can be generalized to both locomotion and manipulation tasks, a significant part of the related literature focuses on locomotion tasks (e.g., \cite{todorov2015ensemble,zimmermann2015hierarchical,mastalli2016hierarchical,manchester2017variational,buchli16_relax,buchli17_mpc}). Nonetheless, for instance, \cite{todorov2012manipulation,todorov2012discovery,posa2014direct,gabiccini2018computational} investigate manipulation tasks but their analysis is either limited to planar case or based on animated characters (i.e., physical realism is not as critical as in robotics).

In manipulation tasks, distances between the robot and objects of interest can be quite large as compared to those in locomotion tasks (i.e., between feet and ground). Hence, smooth contact models are likely to be advantageous for manipulation since they would ease discovering contacts that are initially distant as well as improve convergence. However, existing smooth contact models sacrifice physical accuracy to achieve that. Thus, in this study, we propose a variable smooth contact model that allows the optimizer to adjust the smoothness of the contact model. Moreover, we analyze impacts of different contact models on the performance of contact-implicit trajectory optimization for manipulation. We compare a complementarity constraints-based contact model (CCCM), a smooth contact model (SCM), and the proposed variable smooth contact model (VSCM) for a non-prehensile manipulation task of pushing a cubic object on a tabletop by a Sawyer robot which has a 7 degrees-of-freedom (DOF) arm. We test all methods in simulation for different initial distances between the end effector and the object. The results suggest that the VSCM provides a good trade-off between the physical accuracy and quality of resulting motions at the cost of increased computation time.

The rest of the paper is organized as follows. In Section II, we describe the dynamic model of the system, the contact models, and the corresponding optimization problems. The results and their analyses are presented in Section III. Finally, concluding remarks and the future research directions are given in Section IV.

\section{METHODOLOGY}

\subsection{Dynamic Model}
The dynamics of an $n$ DOF robot that is in contact with environment is given by:
\begin{equation}
    \mathbf{M}(\mathbf{q}) \ddot{\mathbf{q}} + \mathbf{C}(\mathbf{q},\dot{\mathbf{q}}) = \boldsymbol{\tau} + \mathbf{J}_{ext}(\mathbf{q})^T \boldsymbol{\lambda}_{ext},
\end{equation}
where $\mathbf{q}, \dot{\mathbf{q}}, \ddot{\mathbf{q}} \in \mathbb{R}^n $ are the joint positions, velocities, and accelerations, respectively; $\mathbf{M}(\mathbf{q}) \in \mathbb{R}^{n \times n}$ is the inertial matrix; $\mathbf{C}(\mathbf{q},\dot{\mathbf{q}}) \in \mathbb{R}^{n}$ represents the Coriolis, centrifugal, and gravitational terms; $\boldsymbol{\tau} \in \mathbb{R}^n$ is the generalized joint forces; $\mathbf{J}_{ext}(\mathbf{q}) \in \mathbb{R}^{6n_{ext} \times n}$ is the Jacobian matrix mapping the joint velocities to the Cartesian velocities at the external contact points; and $\boldsymbol{\lambda}_{ext} \in \mathbb{R}^{6n_{ext}}$ is the generalized contact forces at the contact points for $n_{ext}$ external contacts. In this study, we use MuJoCo physics engine \cite{mujoco} to model the dynamics.

The vector of generalized joint forces can be decomposed as follows:
\begin{equation}
    \boldsymbol{\tau} = \mathbf{u} + \mathbf{\hat{C}}(\mathbf{q},\dot{\mathbf{q}}) - \mathbf{J}_c(\mathbf{q})^T \boldsymbol{\lambda}_c(\boldsymbol{\gamma}),
\end{equation}
where $\mathbf{\hat{C}}(\mathbf{q},\dot{\mathbf{q}})$ is an estimation of $\mathbf{C}(\mathbf{q},\dot{\mathbf{q}})$ and $\boldsymbol{\lambda}_c\in \mathbb{R}^{6n_c}$ and $\mathbf{J}_c \in \mathbb{R}^{6n_c\times n}$ are respectively the generalized contact forces due to contact model and the corresponding Jacobian matrix  for $n_c$ contact candidates. $\mathbf{u}$ is the control input that is handled by the optimization. $\boldsymbol{\lambda}_c$ is calculated from the magnitude of normal contact force $\gamma\in\mathbb{R}^n_c$ that is determined by either optimization or contact mode. $\boldsymbol{\lambda}_c$ acts on the environment in addition to the forces due to the contact mechanics in MuJoCo (i.e., $\boldsymbol{\lambda}_{ext}$); thus, it can be deemed a virtual force. As a result of this decomposition, the robot can compensate Coriolis, centrifugal, and gravitational effects as well as virtual forces when the control input is zero.

\subsection{Contact Models}
In this study, the following three contact models are investigated. Although the contact models only generate normal contact forces as only pushing manipulation is considered here, MuJoCo takes into account frictional forces in simulations.

\subsubsection{Complementarity Constraints-based Contact Model}
A rigid-body contact model can be formed by the following complementarity constraints:
\begin{align}
    \boldsymbol{\phi}(\mathbf{q}) & \geq \mathbf{0},            \label{eq:cc_phi}   \\
    \boldsymbol{\gamma} & \geq \mathbf{0},                      \label{eq:cc_gamma} \\
    \boldsymbol{\gamma}^T \boldsymbol{\phi}(\mathbf{q}) & = 0,  \label{eq:cc}
\end{align}
where $\boldsymbol{\phi}(\mathbf{q}) : \mathbb{R}^n \rightarrow \mathbb{R}^{n_c}$ is the signed distance function mapping the joint positions to the closest distance between bodies, i.e., (\ref{eq:cc_phi}) prevents interpenetration between bodies. $\boldsymbol{\gamma} \in \mathbb{R}^{n_c}$ is the Lagrange multiplier that physically corresponds to the magnitude of the normal contact force, thus (\ref{eq:cc_gamma}) ensures that bodies can only push each other. Equation~(\ref{eq:cc}), which is evaluated element-wise due to computational reasons (\cite{posa2014direct, fletcher2006local}), guarantees that either distance or contact force must be zero; and these three constraints constitute a complementarity condition that can be denoted by $\mathbf{0} \leq \boldsymbol{\gamma} \perp \boldsymbol{\phi}(\mathbf{q}) \geq \mathbf{0}$.

Recently, Manchester \& Kuindersma \cite{manchester2017variational} proposed a variational contact-implicit trajectory optimization that is similar to \cite{posa2014direct}. However, in \cite{manchester2017variational}, equality constraints in complementarity conditions are relaxed through slack variables to improve the convergence, as in \cite{fletcher2004solving}, by allowing positive contact forces to act from distance while also penalizing this term and thus minimizing deviation from strict rigid-body contact conditions. In this work, we use a similar approach and relax (\ref{eq:cc}) through a slack variable $s \geq 0$ that determines the deviation from the strict rigid-body contact:
\begin{equation}
    \boldsymbol{\gamma}^T \boldsymbol{\phi}(\mathbf{q}) \leq s. \label{eq:relaxed_cc}
\end{equation}
Consequently, (\ref{eq:cc_phi}), (\ref{eq:cc_gamma}), and (\ref{eq:relaxed_cc}) form the complementarity constraints-based contact model that we use in this study.

\subsubsection{Smooth Contact Model}
The relaxation in the complementarity constraints provides a smoother contact model; however, using such a model is restrictive since it requires a constrained optimization algorithm. On the other hand, using directly a smooth contact model, such as those in \cite{marcucci2017two,buchli17_discovery,buchli17_mpc}, would eliminate this limitation. In this study, we use the following exponential formulation for the magnitude of the normal contact force, $\gamma$, that is based on the contact model proposed in \cite{buchli17_mpc} and analogous to a spring model:
\begin{equation}
    \boldsymbol{\gamma}(\mathbf{q}) = k \ e^{(-\alpha \boldsymbol{\phi(\mathbf{q})})} ,
\end{equation}
where $k$ is the spring stiffness and $\alpha$ determines the curvature of the contact force with respect to the distance, $\boldsymbol{\phi}(\mathbf{q})$. We neglect the damping component in the original spring-damper model for the sake of simplicity. Figure \ref{fig:gamma_vs_phi} shows the variation of the magnitude of the normal contact force, $\boldsymbol{\gamma}$, vs. the distance, $\boldsymbol{\phi}(\mathbf{q})$, with negative distance being penetration into the contact surface. 

\subsubsection{Variable Smooth Contact Model}
In the smooth contact model, a fixed value for each parameter is selected such that the convergence of gradient-based optimization is satisfactory. In other words, discovery of contacts is facilitated by allowing, albeit very small, contact forces to act from distance at the cost of loss in physical fidelity. As a consequence, there is a trade-off between the physical accuracy and the convergence of optimization. Thus, the parameters need to be tuned very sensitively based on the task. In order to mitigate this problem, we propose a variable soft contact model in which the spring stiffness $k$ is a decision variable of optimization, and the curvature parameter is a function of $k$, i.e., $\alpha = c/k$ where $c$ is a constant.

\begin{figure}[!tb]
  \centering
  \vspace{0.15cm}
  \includegraphics[width=.5\columnwidth]{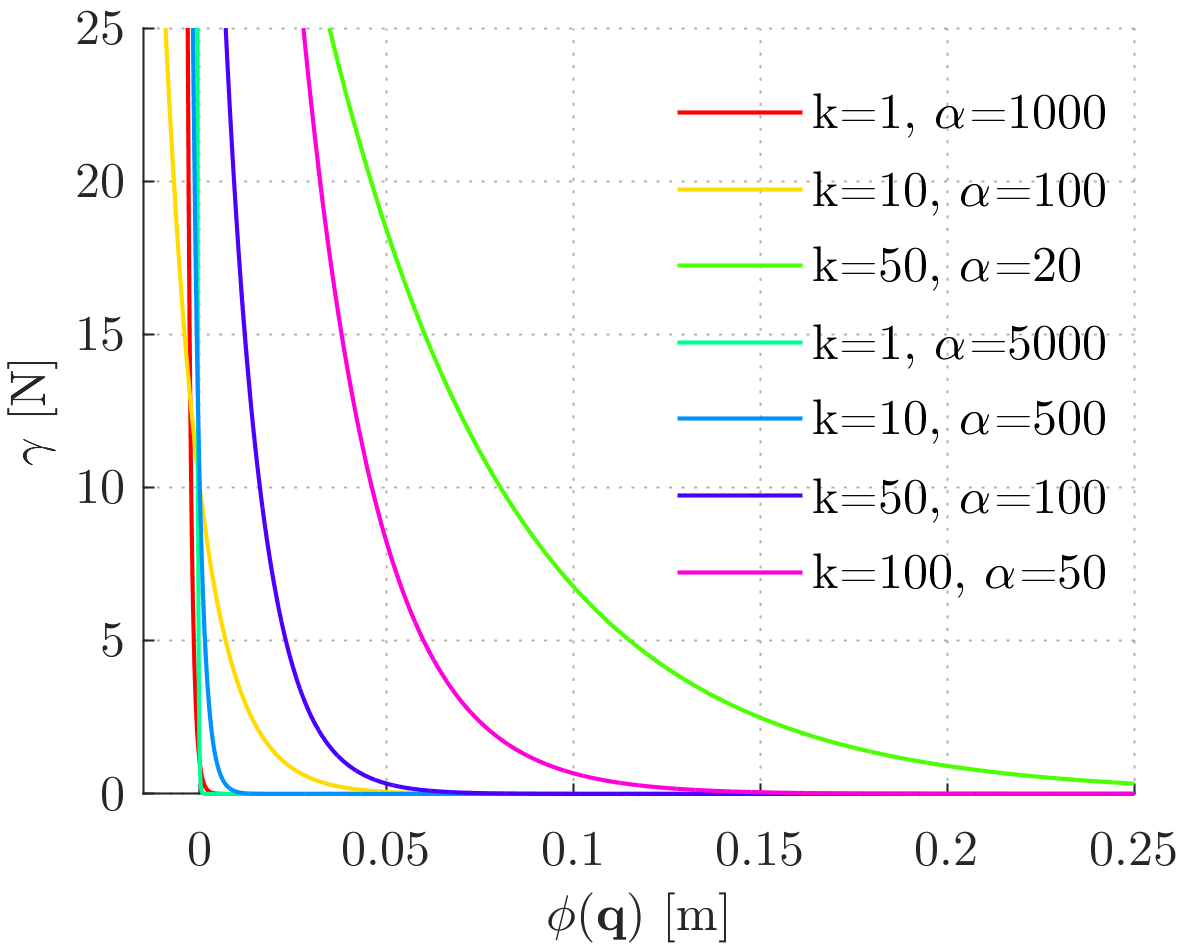}
  \caption{Variation of the magnitude of the normal contact force with respect to the distance for various stiffness and curvature values.}
  \label{fig:gamma_vs_phi}
\end{figure}

In Fig.\ref{fig:gamma_vs_phi}, the variation of $\gamma$ vs. $\phi(\mathbf{q})$ curves with respect to $k$ and $c$ is shown. For the first three curves (indicated by red, yellow, and green), $c=10^3$; and for the remaining curves, $c=5\times10^3$. In this study, we used the greater value for $c$ because the distance - contact force relation converges to a rigid-body contact model as $c$ becomes larger, while it is more similar to a soft contact model for the smaller $c$. It is also noted that the contact force due to this contact model completely vanishes when $k=0$.

Using small values for $k$ is possible when tasks such as locomotion are considered because usually either there are existing contacts or the distances between the contact candidates on the robot and the environment are small. Nevertheless, if one considers a manipulation task that requires, e.g., pushing an object that is far away from the robot, it is necessary to use a much larger $k$ in order for the optimization to be able to discover such motions. However, a large $k$ becomes problematic as the distance goes to zero since making an actual contact with dynamic objects becomes almost impossible. Thus, the VSCM allows the optimization to handle $k$ such that the virtual force $\boldsymbol{\gamma}$ is minimized.

\subsection{Contact-Implicit Trajectory Optimization}
In the contact-implicit trajectory optimization, the main goal is to find contact forces along with control inputs given a high-level definition of the task in terms of an optimization problem. The optimal control problem of trajectory optimization (i.e., an infinite-dimensional problem) can be converted into a finite-dimensional problem through transcription methods which can be divided into three categories as single shooting, multiple shooting, and direct transcription \cite{betts1998survey}. In this study, we transcribe the problem into a single-shooting optimization problem by using evenly separated collocation points for the control inputs.

We consider the task of non-prehensile manipulation of pushing a cubic object on a tabletop by a 7 DOF robotic arm. The corresponding cost function can be written in terms of final costs and integrated costs. In this case, the final costs are based on the distance of the object's center of mass (CoM) from a desired position, $\mathbf{p}_{o}^e$, and the deviation of the object's orientation from desired Euler angles, $\boldsymbol{\theta}_o^e$; while the integrated costs are based on the end-effector velocity, $\dot{\mathbf{x}}(\mathbf{q},\dot{\mathbf{q}})$, and the magnitude of the normal contact force, $\boldsymbol{\gamma}$. Moreover, we normalize these cost terms such that $\mathbf{p}_{o}^e$ is divided by the norm of the initial position error, which yields $\mathbf{p}_{o,normalized}^e$, and the integrated costs are divided by $N$ so that the optimization process is less sensitive to the task and the duration. As a result, the final and integrated components of the task-related cost ($c^f$ and $c^i$, respectively) are calculated in terms of the weights $w_{1,...,4}$, the control sampling period $t_c$, the time step $l$, and the number of time steps $N$ by:
\begin{align}
    c^f = w_1  ||\mathbf{p}_{o,normalized}^e||_2^2 + w_2 ||\theta_o^e||_2^2, \\
    c^i = \frac{t_c}{N} \sum_{l=1}^N (w_3||\dot{\mathbf{x}}_l(\mathbf{q}_l,\dot{\mathbf{q}_l})||_2^2 +w_4 ||\boldsymbol{\gamma}_l||_2^2).
\end{align}
The following optimization problems are solved by a sequential quadratic programming (SQP) algorithm by running forward simulations in MuJoCo to evaluate cost and constraint functions.

\subsubsection{Optimization Problem for CCCM}
The optimization problem that includes the relaxed complementarity constraints is defined as:
\begin{equation}
    \underset{\mathbf{u}_{1,...,N},\mathbf{\gamma}_{1,...,N},s_{1,...,N}}{\text{minimize }} c^f + c^i + w_5 \sum_{l=1}^N s_l^2
\end{equation}
subject to
\begin{align}
    \boldsymbol{\phi}_l(\mathbf{q}), \mathbf{\gamma}_l, s_l \geq \mathbf{0} \text{ for } l = 1,...,N \\
    \boldsymbol{\phi}_l^T(\mathbf{q}) \mathbf{\gamma}_l \leq s_l \text{ for } l = 1,...,N
\end{align}
For this problem, the weights are selected as $w_1=10^4$, $w_2=10^4$, $w_3=2 \times 10^{-1}$, $w_4=2 \times 10^{-1}$, and $w_6=10^3$.

\subsubsection{Optimization Problem for Smooth Contact Models}
The optimization problem takes the following unconstrained form for smooth contact models:
\begin{equation}
    \underset{\mathbf{u}_{1,...,N}, (k_{1,...,N})}{\text{minimize }} c^f + c^i.
\end{equation}
The only difference is that the stiffness for each control time step $k_l$ is a decision variable for the variable SCM. The weights for both problems are same and selected as $w_1=10^3$, $w_2=10^3$, $w_3=2 \times 10^{-1}$, and $w_4=2 \times 10^{-2}$.

\section{Simulation Experiments}
\subsection{Experimental Setup}
The algorithms are tested in simulation for a simple scenario of pushing a cubic object on a tabletop with a Sawyer robot that has a 7 DOF arm. For simulations, we use MuJoCo physics engine since it is shown to be favorable for robotic systems with contacts \cite{mujoco_vs_solvers}. Optimization problems are solved through an SQP-based solver called SNOPT \cite{snopt}. Additionally, we use IFOPT \cite{ifopt} as an interface between MuJoCo and SNOPT.

\begin{figure}[!tb]
  \centering
  \vspace{0.15cm}
  \includegraphics[width=0.45\columnwidth]{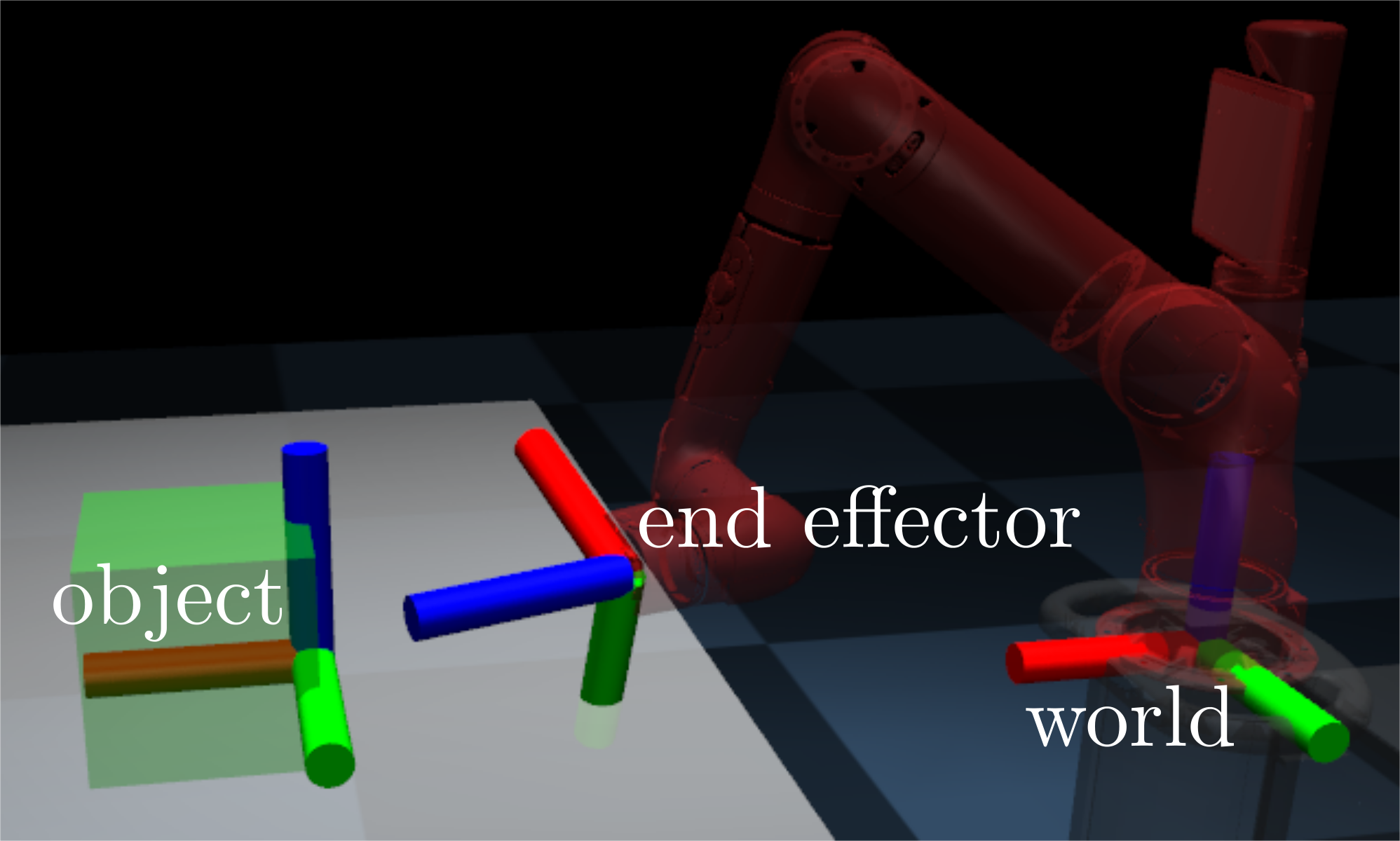}
  \caption{Simulation environment along with the world frame and the frames associated with the contact candidates on the robot and the object.}
  \label{fig:sim_env_sites}
\end{figure}

In order to keep the complexity of the task low for the sake of comparison, we consider a simple pushing task in which the robot needs to push a cubic box 0.25 m along the $x$-axis of the world frame (see Fig. \ref{fig:sim_env_sites}) without rotating it. Therefore, we take into account only two contact candidates: one on the object's closest surface to the robot, and one at the center of the end-effector plate of the robot. The simulation environment and the contact frames associated with the contact candidates on the robot and the object are depicted in Fig. \ref{fig:sim_env_sites}. Nonetheless, it is noteworthy that the robot can make contact at arbitrary positions on its surface.

For all simulations, the duration is 1 s, and the control sampling period $t_c=50$ ms, i.e., $N=20$. The initial configuration of the arm is calculated by the following function of $\beta$ that also determines the distance between the contact candidates $\mathbf{q}_0 = [-\pi/12, -\pi/\beta, 0, 3\pi/\beta, -\pi/6, -2\pi/\beta, 0]^T$.

\begin{figure}[!tb]
  \centering
  \subfigure[$\beta=5, \ \phi_0=0.30$ m]{\includegraphics[width=0.35\columnwidth]{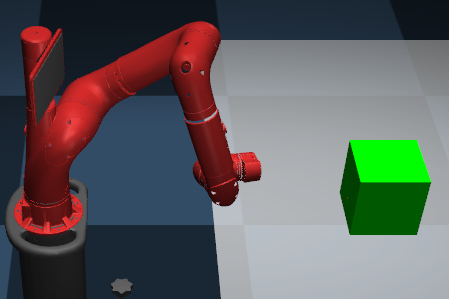}}
  \subfigure[$\beta=6, \ \phi_0=0.17$ m]{\includegraphics[width=0.35\columnwidth]{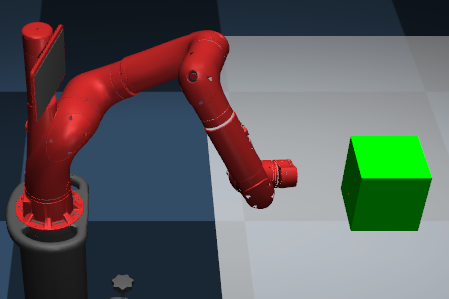}}
  \subfigure[$\beta=7, \ \phi_0=0.11$ m]{\includegraphics[width=0.35\columnwidth]{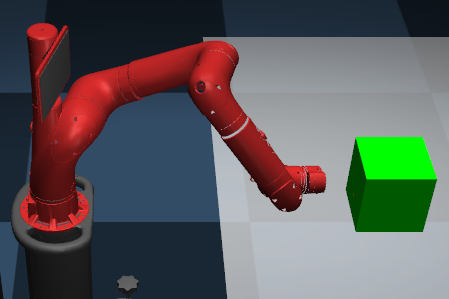}}
  \caption{The initial configurations of the robot for different values of $\beta$.}
  \label{fig:init_configs}
\end{figure}

We run simulations for three different values of $\beta$ that are 5, 6, and 7, and the corresponding initial distances between the contact candidates are 0.30 m, 0.17 m, and 0.11 m, respectively. Figure \ref{fig:init_configs} demonstrates the three initial configurations of the arm. It is seen that not only the initial position but also, albeit slightly, the orientation of the end effector vary depending on $\beta$. The purpose of different initial configurations is to change the complexity of the task in a controlled manner by assuming that difficulty in discovering contacts augments as the distance increases.

The stiffness parameter of the SCM is selected as 100 after a careful tuning considering the overall performance, and the corresponding $\alpha$ is calculated by $c=5\times10^3/k$, as in the VSCM. We use the same value to initialize and upper bound $k$ in the VSCM-based optimization (VSCMO). This value seems reasonable when one looks at the distance range here and the $\gamma$ vs. $\phi(\mathbf{q})$ curves in Fig. \ref{fig:gamma_vs_phi}. In all cases, the initial guess for $u$ is zero, namely the robot is stationary throughout the simulation.

\subsection{Visual Analysis}
\begin{figure}[!tb]
  \centering
%   \vspace{0.10cm}
  \subfigure{\includegraphics[width=0.43\columnwidth]{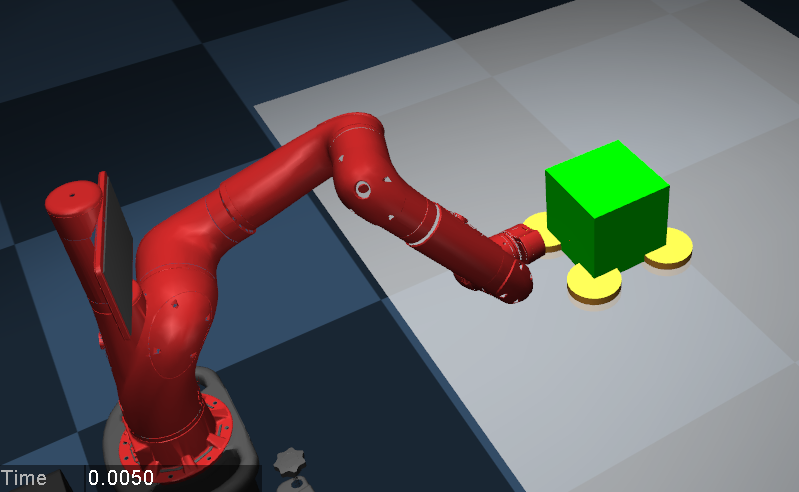}}
  \subfigure{\includegraphics[width=0.43\columnwidth]{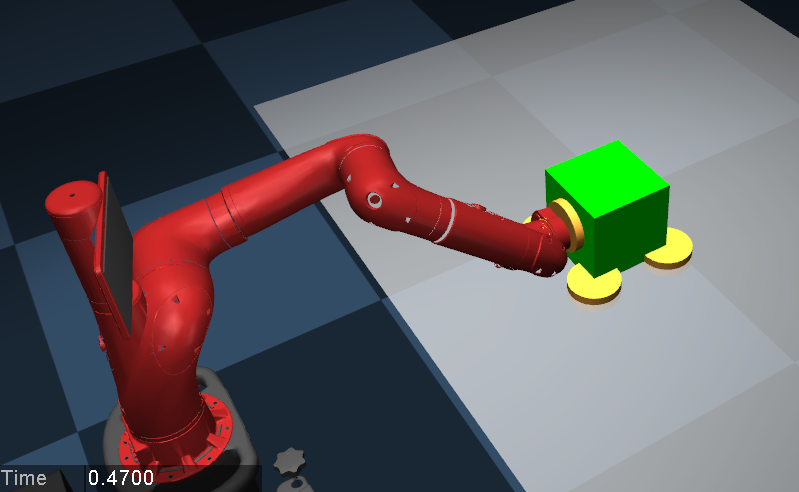}}
  \subfigure{\includegraphics[width=0.43\columnwidth]{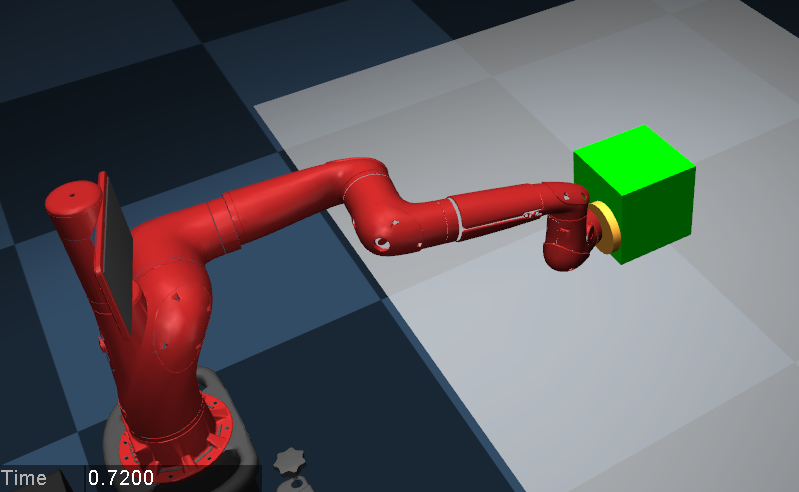}}
  \subfigure{\includegraphics[width=0.43\columnwidth]{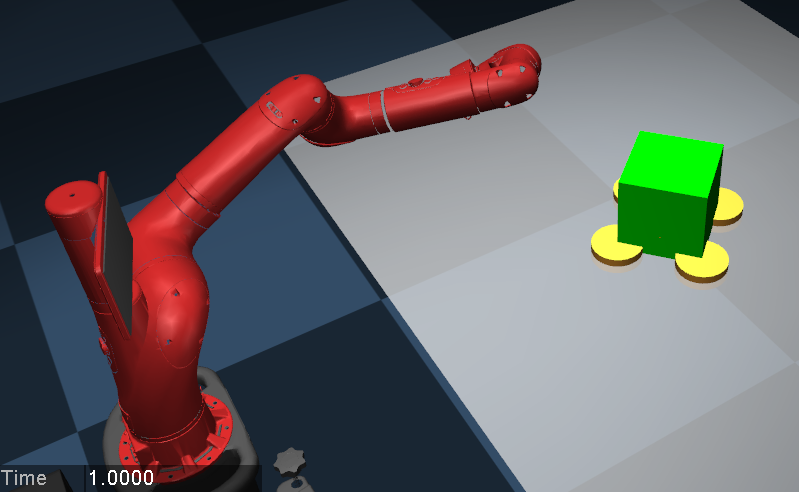}}
  \caption{Snapshots of a resulting motion obtained for the CCCM for t = 5, 470, 720, and 1000 ms. Yellow cylinders represent active contacts.}
  \label{fig:arm7_cccm}
\end{figure}

In order to observe the discrepancies between motions that are optimized by assuming different contact models, we visualize the resulting motions in the following.\footnote{Please see the accompanying video for all simulation results: \url{https://youtu.be/u06ZTmut0N0}.} First, a motion found by the CCCM-based optimization (CCCMO) is shown in Fig. \ref{fig:arm7_cccm}. In this motion, the first contact between the robot and the object is made at t=470 ms. Then, the contact starts to slide in the $-y$~direction by causing an undesired rotation of the object in addition to the desired translation. After the contact is broken, the arm rises due to the impact.

\begin{figure}[!tb]
  \centering
  \subfigure{\includegraphics[width=0.43\columnwidth]{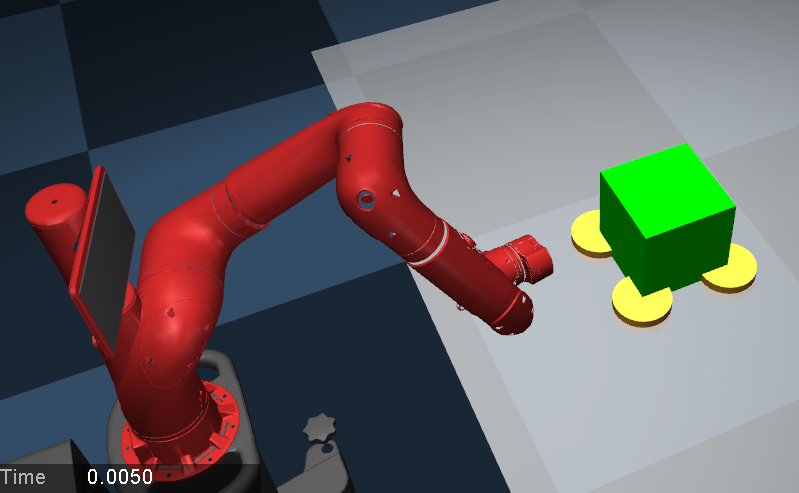}}
  \subfigure{\includegraphics[width=0.43\columnwidth]{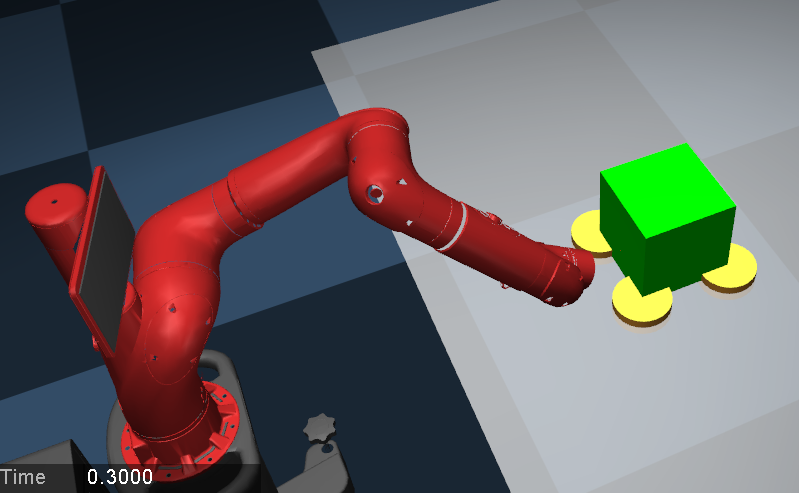}}
  \subfigure{\includegraphics[width=0.43\columnwidth]{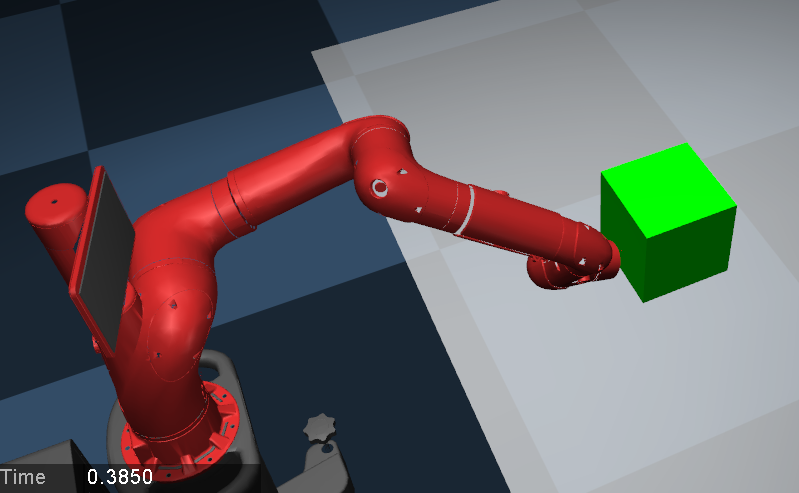}}
  \subfigure{\includegraphics[width=0.43\columnwidth]{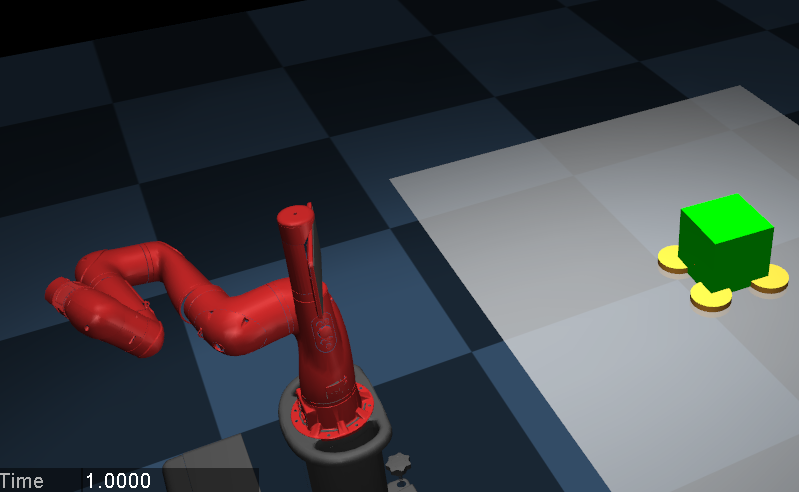}}
  \caption{Snapshots of a resulting motion obtained for the SCM for t = 5, 300, 385, and 1000 ms. Yellow cylinders represent active contacts.}
  \label{fig:arm6_scm}
\end{figure}

Second, Fig. \ref{fig:arm6_scm} illustrates a resulting motion of the SCM-based optimization (SCMO). The robot approaches the object, and the object is still stationary at t=300 ms. However, at t=385 ms, the object starts to move due to the force generated by the SCM, even though there is no actual contact between the robot and the object. At the end, the object is successfully manipulated with no visible rotation; yet the arm is repelled significantly due to the large $\gamma$ values during pushing. The problem of pushing without an actual contact might be alleviated by a very sensitive tuning of $k$; however, an actual contact still would be very hard to achieve in conjunction with satisfactory motion and convergence. Moreover, repeated sensitive tuning is likely to be necessary even for slight changes in the task, such as the initial configuration of the robot.

\begin{figure}[!tb]
  \centering
%   \vspace{0.15cm}
  \subfigure{\includegraphics[width=0.43\columnwidth]{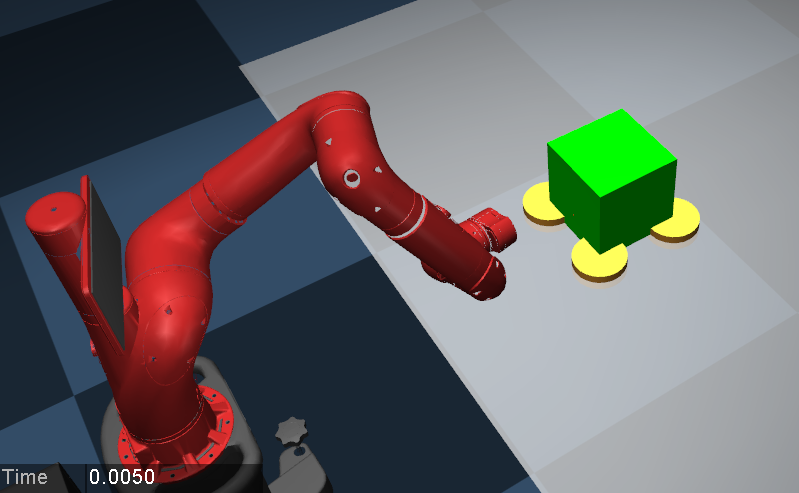}}
  \subfigure{\includegraphics[width=0.43\columnwidth]{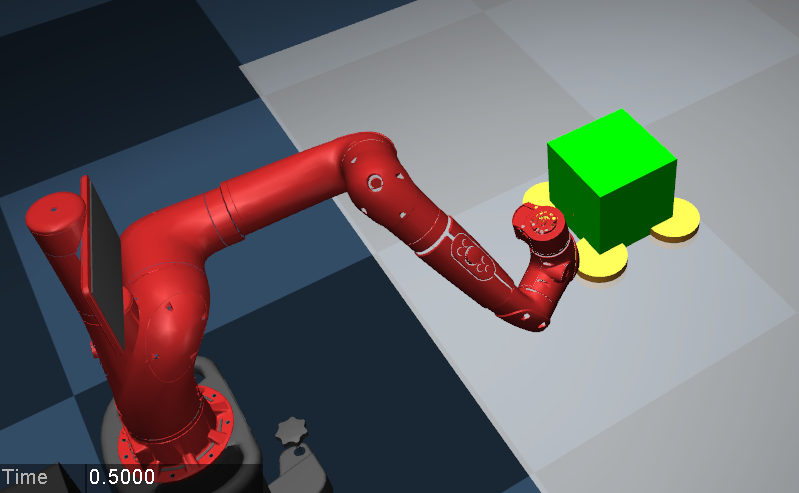}}
  \subfigure{\includegraphics[width=0.43\columnwidth]{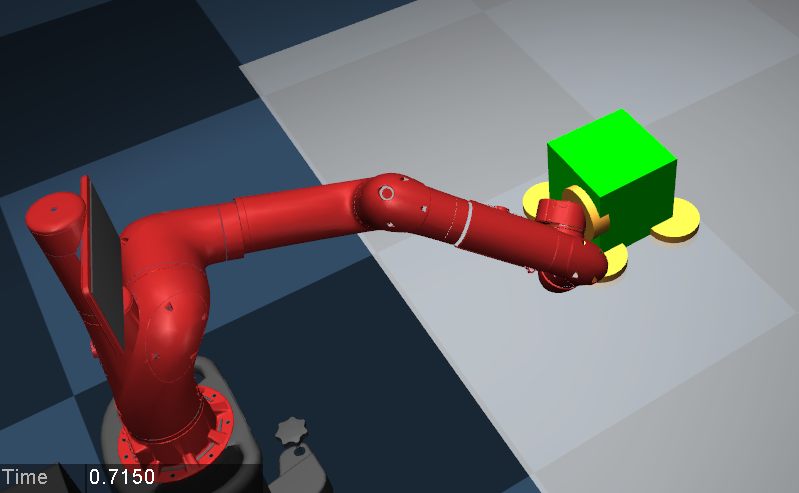}}
  \subfigure{\includegraphics[width=0.43\columnwidth]{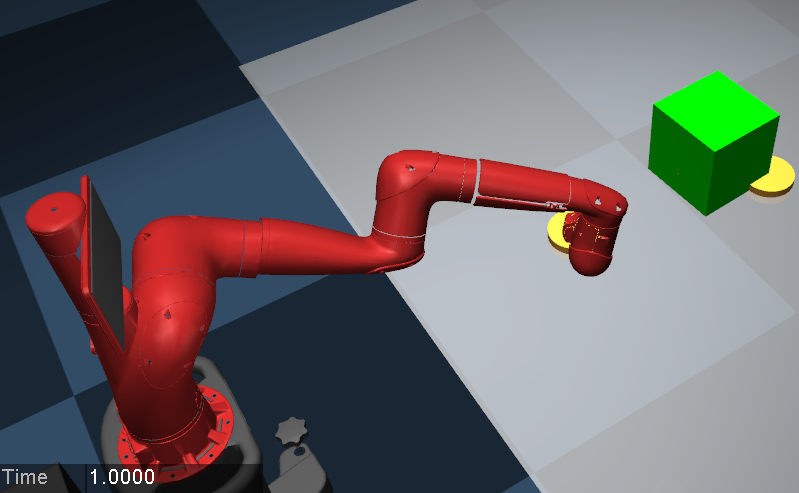}}
  \caption{Snapshots of a resulting motion obtained for the VSCM for t = 5, 500, 715, and 1000 ms. Yellow cylinders represent active contacts.}
  \label{fig:arm6_vscm}
\end{figure}

Last, a motion obtained from the VSCMO is demonstrated in Fig. \ref{fig:arm6_vscm}. Here, the object does not move until there is an active contact at t=715 ms. Then, the object is pushed successfully to the desired pose, while the arm is not repelled significantly and stops by making contact with the table. In this case, the arm actually contacts the object. This is possible owing to the adjustment of stiffness parameter $k$ by the optimization.

In addition to these, it is clearly seen that the contact at t=715 ms occurs at neither of the contact candidates on the robot nor the object. This is possible due to the fact that the robot can make and break contacts at arbitrary parts of its surface by the virtue of MuJoCo physics engine.

\subsection{Numerical Analysis}
After the visual analysis, we investigate the numerical results obtained from the simulations for all the contact models and the initial configurations in the following. First, we focus on the physical fidelity of the resulting motions. In order to measure the physical inaccuracy throughout a simulation, we use a metric that is calculated by integrating the magnitude of the normal contact force acting from distance, i.e., $t_c \sum_{k=1}^N \gamma_k$. Table \ref{tab:phys_inacc} shows the values of this metric for all cases. It is seen that the CCCMO provides the best physical fidelity by far for the first two cases. Whereas, the VSCMO has the minimum physical inaccuracy in the last case, namely there is almost no virtual force. Moreover, it provides a much better physical accuracy over the SCMO in all cases.

\begin{table}[ht]
    \caption{Physical Inaccuracy Metric}
    \begin{center}
        \begin{tabular}{|c||c|c|c|}
            \hline
            $\phi_0$ [m] & CCCMO [N-s] & SCMO [N-s] & VSCMO [N-s] \\
            \hline
            0.11         & 0.1121     & 2.0400    & 1.2137 \\
            \hline
            0.17         & 0.0238     & 1.4914    & 1.2676 \\
            \hline
            0.30         & 0.0373     & 0.0381    & 0.0001 \\
            \hline
        \end{tabular}
    \end{center}
  \label{tab:phys_inacc}
\end{table}

Tables \ref{tab:pos_err} and \ref{tab:orient_err} demonstrate the final position and orientation errors for all cases, respectively. It is seen that in all cases, the VSCMO provides the best performance in terms of the manipulation task, when it is assumed that the differences between the orientation errors for the SCM and VSCM in the first two cases are negligible. Furthermore, it is the only method that can perform the task satisfactorily (i.e., with a position error smaller than 10 cm) for all initial configurations. On the other hand, the SCMO is also able to perform the task successfully except for the last case; while the CCCMO can find an \textit{almost} successful motion only for the smallest initial distance, and even in that case, the orientation error is quite large.

\begin{table}[ht]
    \vspace{0.15cm}
    \caption{Final Position Error}
    \begin{center}
        \begin{tabular}{|c||c|c|c|}
            \hline
            $\phi_0$ [m] & CCCMO [m]   & SCMO [m]   & VSCMO [m] \\
            \hline
            0.11         & 0.1052     & 0.0285    & 0.0185 \\
            \hline
            0.17         & 0.1979     & 0.0318    & 0.0291 \\
            \hline
            0.30         & 0.1924     & 0.1908    & 0.0836 \\
            \hline
        \end{tabular}
    \end{center}
  \label{tab:pos_err}
\end{table}

\begin{table}[ht]
    \caption{Final Orientation Error}
    \begin{center}
        \begin{tabular}{|c||c|c|c|}
            \hline
            $\phi_0$ [m] & CCCMO [rad] & SCMO [rad] & VSCMO [rad] \\
            \hline
            0.11         & 0.7067     & 0.0238    & 0.0533 \\
            \hline
            0.17         & 0.4022     & 0.0586    & 0.0654 \\
            \hline
            0.30         & 0.3475     & 0.4547    & 0.0415 \\
            \hline
        \end{tabular}
    \end{center}
  \label{tab:orient_err}
\end{table}

The computation times for all cases are shown in Table \ref{tab:comp_time}. We admit that our implementation is significantly slower than the state-of-the-art counterparts such as \cite{buchli17_discovery} and, in particular, \cite{buchli17_mpc}. However, an optimized implementation of the presented methods is out of scope of this work. Here, we are rather interested in comparing them in terms of the convergence speed. Based on these results, it is not possible to say that there is a clear relationship between the initial distance (or the complexity of the task) and the computation time; i.e., there is a positive correlation for the CCCM, a negative correlation for the SCM, and no correlation for the VSCM. However, it is fair to say that, excluding the second case, the SCMO converges significantly faster than the VSCMO and has the fastest convergence, as anticipated. On the other hand, the CCCMO has the slowest convergence speed except for the first case in which its computation time is slightly larger than SCM.

\begin{table}[ht]
    \caption{Computation Time}
    \begin{center}
        \begin{tabular}{|c||c|c|c|}
            \hline
            $\phi_0$ [m] & CCCMO [s] & SCMO [s]   & VSCMO [s] \\
            \hline
            0.11         & 43.09    & 41.07     & 111.22 \\
            \hline
            0.17         & 44.52    & 27.87     & 16.28 \\
            \hline
            0.30         & 112.77   & 23.98     & 60.49 \\
            \hline
        \end{tabular}
    \end{center}
  \label{tab:comp_time}
\end{table}

\subsection{Variation of Stiffness for VSCMO}
\begin{figure}[!tb]
  \centering
  \vspace{0.15cm}
  \includegraphics[width=0.65\columnwidth]{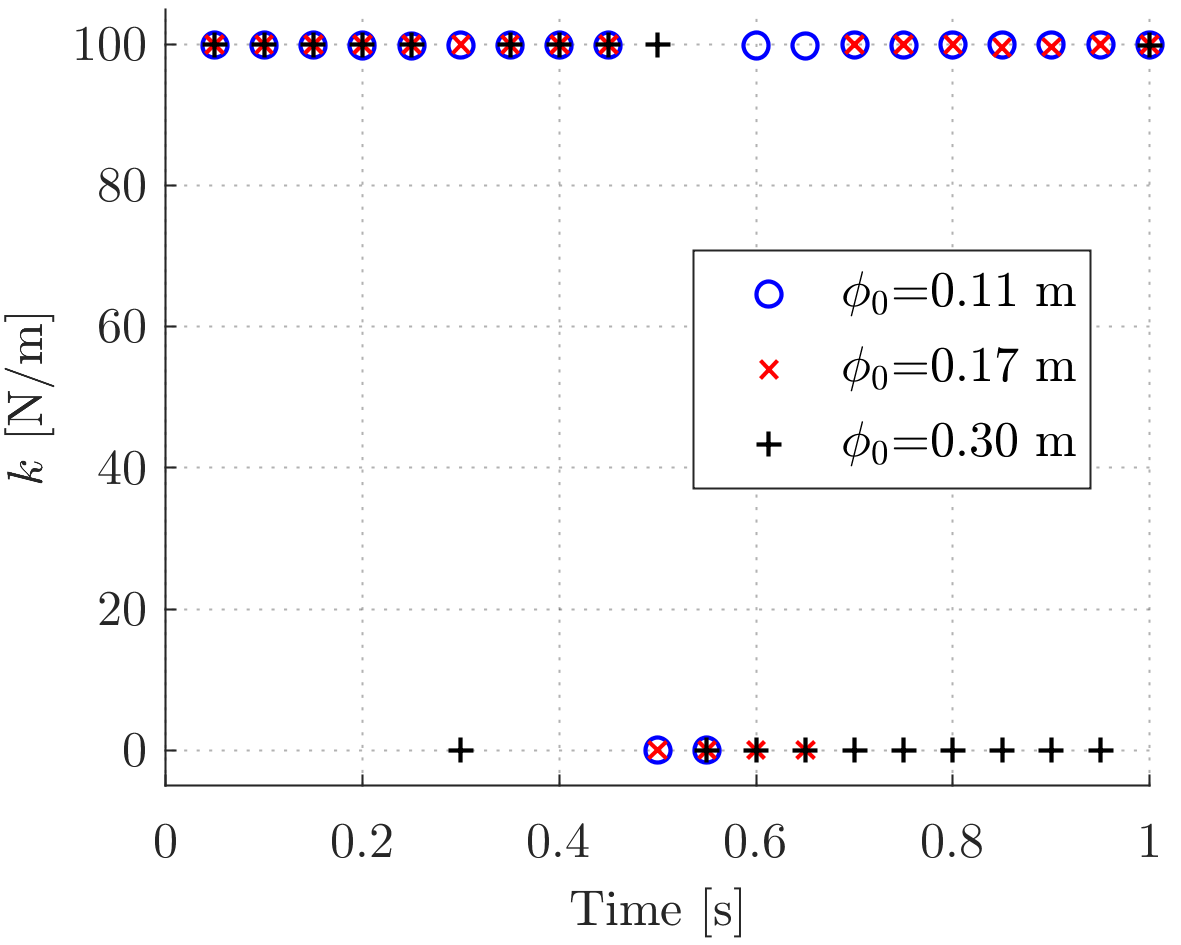}
  \caption{$k$ trajectories found by the VSCM-based optimization for the three values of the initial distance between the contact candidates.}
  \label{fig:ks_t}
\end{figure}

Finally, the $k$ trajectories obtained from the VSCMO are depicted in Fig. \ref{fig:ks_t} for all initial configurations.\footnote{Please see the accompanying video for the virtual and actual contact force trajectories for all cases.} It is seen that the optimization handles the stiffness $k$ as if it is a binary variable; i.e., $k$ is either around 100, its initial value and the upper bound, or equal to zero. The collocation points when it is equal to zero correspond to the instants of the simulation when the arm is approaching to the object. This is caused by the fact that the virtual force generated by the contact model, $\boldsymbol{\gamma}$, that is proportional to $k$ and inversely proportional to the distance, $\boldsymbol{\phi}$, is penalized. Consequently, in the VSCMO, contacts can be easily discovered from distance as in the SCMO as well as the robot can actually make contact with the object to physically move it, as distinct from the case for the SCM with fixed parameters (see Figs. \ref{fig:arm6_scm} and \ref{fig:arm6_vscm}). Moreover, this can be achieved for a wide range of tasks without a sensitive tuning of the contact model parameters.

\section{CONCLUSION}

In this paper, we analyze the impact of contact models on the performance of contact-implicit trajectory optimization for manipulation. The trajectory planning method in question is a single shooting optimization. Three different contact modeling approaches are investigated: (1) a complementarity constraints-based contact model, (2) a smooth contact model, and (3) our proposed variable smooth contact model. We perform simulations of a 7 DOF Sawyer robot arm carrying out a pushing manipulation task of varying complexity. 

An analysis of the resulting trajectories yields the following observations. First, all of the methods are capable of discovering a pushing motion given zero initial values for the torque variables. Second, the proposed approach, VSCMO, is the most reliable as it finds satisfactory motions for all initial configurations. Third, the VSCMO provides the best performance when both the quality and the physical accuracy of the motions found are considered. Notwithstanding, its convergence is slower than the SCMO which outperforms the CCCMO in all cases. However, the SCMO is very sensitive to the parameters of the SCM, and therefore, it is hard to tune for a range of tasks. On the other hand, the VSCM allows the optimization to vary the smoothness of the contact model as necessary. Consequently, the VSCMO is more robust to changes in the task and, thus, more suitable for contact-implicit trajectory optimization for manipulation. 

Nevertheless, the task we investigate here in this paper is limited to one contact candidate on both the robot and the object. We aim to test the proposed method on more complex tasks and behaviours. Furthermore, in future work, we aim to implement our proposed method experimentally.

\addtolength{\textheight}{-0.12cm}   % This command serves to balance the column lengths

%%%%%%%%%%%%%%%%%%%%%%%%%%%%%%%%%%%%%%%%%%%%%%%%%%%%%%%%%%%%%%%%%%%%%%%%%%%%%%%%

% %% Use plainnat to work nicely with natbib. 
% \bibliographystyle{IEEEtran}

% % \bibliographystyle{plainnat}
% \bibliography{refs}


\end{document}